\documentclass[submission]{article} % DO NOT CHANGE THIS
\usepackage{aaai25}  % DO NOT CHANGE THIS
\usepackage{times}  % DO NOT CHANGE THIS
\usepackage{helvet}  % DO NOT CHANGE THIS
\usepackage{courier}  % DO NOT CHANGE THIS
\usepackage[hyphens]{url}  % DO NOT CHANGE THIS
\usepackage{graphicx} % DO NOT CHANGE THIS
\usepackage{natbib}  % DO NOT CHANGE THIS AND DO NOT ADD ANY OPTIONS TO IT
\usepackage{caption} % DO NOT CHANGE THIS AND DO NOT ADD ANY OPTIONS TO IT
\usepackage{newfloat}
\usepackage{listings}
\usepackage{amssymb}
\usepackage{multirow}
\usepackage{diagbox}
\usepackage[norelsize, linesnumbered, ruled, lined, boxed, noend]{algorithm2e} % For algorithms
\usepackage{floatrow}
\usepackage[dvipsnames]{xcolor}
\definecolor{Green4}{RGB}{0,128,0}
\usepackage{newfloat}
\usepackage{listings}
\usepackage{subcaption}

\DeclareCaptionStyle{ruled}{labelfont=normalfont,labelsep=colon,strut=off} % DO NOT CHANGE THIS
\floatstyle{ruled}
\newfloat{listing}{tb}{lst}{}
\floatname{listing}{Listing}
\title{State-Based Disassembly Planning}
\author {
    Chao Lei, Nir Lipovetzky, Krista A. Ehinger \\
}
\affiliations {
    School of Computing and Information Systems, The University of Melbourne, Australia\\
    clei1@student.unimelb.edu.au, \{nir.lipovetzky, kris.ehinger\}@unimelb.edu.au

}

\usepackage{bibentry}
\begin{document}

\maketitle

\begin{abstract}
It has been shown recently that physics-based simulation significantly enhances the disassembly capabilities of real-world assemblies with diverse 3D shapes and stringent motion constraints. However, the efficiency suffers when tackling intricate disassembly tasks that require numerous simulations and increased simulation time. 
In this work, we propose a State-Based Disassembly Planning (SBDP) approach, prioritizing physics-based simulation with translational motion over rotational motion to facilitate autonomy, reducing dependency on human input, while storing intermediate motion states to improve search scalability. We introduce two novel evaluation functions derived from new \textit{Directional Blocking Graphs} (DBGs) enriched with state information to scale up the search.  Our experiments show that SBDP with new evaluation functions and DBGs constraints outperforms the state-of-the-art in disassembly planning in terms of success rate and computational efficiency over benchmark datasets consisting of thousands of physically valid industrial assemblies. 
\end{abstract}

\section{Introduction}
 
Assembly planning is crucial for optimizing manufacturing efficiency and minimizing maintenance and repair costs \cite{ghandi2015review}. However, manual assembly planning, primarily developed by experienced engineers, will become more difficult in the era of industry 4.0, where product manufacturing transits from mass production to mass customization \cite{froschauer2021human}. Numerous methods for automation have been proposed \cite{perrard2023new,bedeoui2018assembly,tian2022assemble}  through various methodologies, such as heuristic search \cite{wang2013enhanced}, evolutionary algorithms \cite{zeng2011multi} and neural networks \cite{chen2008three}. Despite these efforts, automatically generating an efficient, precise and generalizable assembly plan remains an open challenge.

An assembly planning task often contains two steps: assembly sequence planning, which computes the sequential order to assemble all components; and assembly path planning, which searches for penetration-free trajectories to add new components into a sub-assembly. The bijection between assembly and disassembly sequences, when all parts are rigid, raises the idea of assembly-by-disassembly, introduced by \citet{demello1991}. The assembly-by-disassembly method, understood as regression, minimizes the required search space in assembly tasks by leveraging the defined precedence and motion constraints in assembled parts. It has been widely implemented in different assembly tasks, including mechanical product assembly \cite{demello1991, wang2014mechanical} and kit assembly \cite{zakka2020form2fit}. 

With the development of general-purpose assembly planning algorithms,  \citet{tian2022assemble} proposed a physics-based assembly-by-disassembly planning method, referred as PDP, for \textit{sequential} assembly tasks where  operations involve inserting an individual part into the sub-assembly.
This approach employs a physics-based simulation,  built upon the rigid body simulator developed by \citet{xu2021end} to generate the disassembly trajectories while using a sequential planner to determine the disassembly sequence. The disassembly trajectories and sequence are reversed to construct the assembly plan. 
Compared to sampling-based approaches such as RRT \cite{lavalle1998rapidly} and its variants  \cite{aguinaga2008parallel, ebinger2018general}, as well as the classic physics-based method \cite{zickler2009efficient}, PDP demonstrates state-of-the-art performance in terms of success rate and computational efficiency.

Despite its strong performance, PDP suffers from low efficiency since the exhaustive search in disassembly sequence planning and path planning results in redundant and unnecessary simulations. Regenerating already simulated trajectories instead of reusing them also weakens the efficiency of PDP.
In addition, users must specify if a task requires translational or rotational operations, as considering both at the same time hinders the advantages of using one operation at a time, where translational motion offers computational efficiency, while rotational motion provides better coverage. Requiring user input reduces the autonomy of the deployed solutions in PDP. To address these issues, we prioritize translational motion over rotational motion in a State-Based Disassembly Planning search algorithm. We use the physical simulation as a transition function and record the trajectories in the state representation, avoiding computationally intensive simulations of states that need to be revisited. We integrate and propose novel \textit{Directional Blocking Graphs} with state information to derive new evaluation functions that allow disassembly planning to accurately select components and actions for disassembly, reducing further the overall search space.

\section{Background}

A common strategy for solving planning problems is to create a state model for the target domain \cite{bonet2001planning}. Formally, a state model is a tuple $\mathcal
{S} = \langle{S, s_{0}, S_{G}, A, f, c}\rangle$ where $\mathcal{S}$ consists of  a set of states $S$, a set of actions $A$, where $A(s) \subseteq A$ are applicable actions in each state $s \in S$, an initial state  $s_0 \in S$, a set of goal states $S_G \subseteq S$, a deterministic transition function $s^\prime=f(a, s)$  that defines how action $a\in A(s)$ map one state $s$ into another state $s^\prime$, and the cost function $c(a, s)$ that estimates the cost of applying action $a$ in state $s$. The state model is well-suited for the disassembly planning task given that actions are defined and disassembly configurations are fully observed.

We formulate a \textit{sequential} 
%\textit{linear monotone two-handed} 
disassembly problem with a state model. A disassembly problem $\mathcal{P}$ consisting of $m$ parts $P=\{p_1,\ldots,p_m\}$ is described by a tuple $\mathcal{P}=\langle P, S, A, s_0, S_G, f \rangle$,  where $S=(\vec{s}_1,\ldots,\vec{s}_m)^T \in \mathbb{R}^{3\times m}$ describes the translational vector of each part for translational motion problems with actions $A=\{(\pm 1,0,0),(0,\pm 1,0),(0,0,\pm 1)\}$; otherwise, $S$$=SE(3)^{m}$ describes the transformation matrix of each part with actions $A=\{(\pm 1,0,0,0,0,0),\ldots, (0,0,0,0,0, \pm 1)\}$, corresponding to 6 translational and 6 rotational degrees of freedom, for rotational motion problems. Each action specifies a positive or negative change in one dimension for a single movable part $p \in P$. $s_0$ denotes the initial assembled state of all parts $s_0\in S$. $s^\prime=f(a, s, p)$ is the state transition function implemented by the physical simulation. The physical simulation generates a penetration-free motion trajectory (path) from $s$ to $s^\prime$, denoted as $t(s, s^\prime, a, p)$, by continuously moving $p$ from state $s$ with action $a \in A$, and ultimately returning a state $s^\prime$.  We call $s^\prime$ as a collision state  if $p$ at $s^\prime$ has a collision with any other components, or a disassembled state that satisfies the geometry constraint, iff the convex hull of the geometry of part $p$ at $s^\prime$ does not intersect with the convex hull encompassing the geometries of all other components $P\setminus p$ at $s^\prime$. $S_G\subseteq S$ specifies the set of disassembled goal states defined as the states that satisfy the geometry constraint for every part in $P$. 
A partial disassembly path $\mathcal{T}_{p}=\{t(s_0, s_1, a_0, p) \ldots,t(s_{n-1}, s_n, a_{n-1}, p)\}$, consisting of a sequence of motion paths for a single part $p$, is a disassembly path when $s_n$ is the disassembled state for $p$. A disassembly plan, $\mathcal{DP}=\{\mathcal{T}_{p_1},\ldots,\mathcal{T}_{p_m}\}$, comprises a sequence of ordered disassembly paths for every part in $P$. 

We note that once the translational vector or transformation matrix encoded in a state is provided, the specific location can be determined by applying it to the original coordinates. In other words, the state $s=\{s_{p_1}, \ldots, s_{p_m}\}$ describes the location of each part, where we use $s_{p_m}$ to represent the location of $p_m$ at state $s$. $s$ is the initial state $s_0$ when the location of each part in $s$ is the initial location in $\mathcal{P}$. $s$ is a goal state $s_g$ when each part at its location in $s$ meets the geometry constraint. The transition function $s^\prime=f(a, s, p_1)$ updates the state $s=\{s_{p_1}, \ldots, s_{p_m}\}$ to the state $s^\prime=\{s^\prime_{p_1}, \ldots, s^\prime_{p_m}\}$ through the physics-based simulation, where only $s^\prime_{p_1}$ is changed while $\{s^\prime_{p_2}, \ldots, s^\prime_{p_m}\}=\{s_{p_2}, \ldots, s_{p_m}\}$ as only $p_1$ is simulated. Subsequently, only $p_1$ is available  to simulate in state $s^\prime$ since we only consider the sequential disassembly problem.

To keep the simulation computationally manageable, we follow the same assumptions as introduced by \citet{tian2022assemble}: 1) assemblies consist entirely of rigid components;  2) gravity, friction and manipulation constraints are omitted;  3) parts can be fully assembled or disassembled sequentially.

\subsection{Physics-Based Disassembly Planning} 

PDP leverages a breadth-first search (BFS) as the path planner to disassemble one part at a time. In the search tree, each node represents the location configuration of an assembled part, and each edge corresponds to the physical simulation with an action that updates the location of an assembled part, resulting in a child node. Given an assembled part $p$,  the path planner searches for a sequence of actions until a disassembled state has been found or a depth limit on the longest sequence has been reached. If a  part $p$ is successfully disassembled, its disassembly path $\mathcal{T}_{p}$ is attached to the disassembly plan $\mathcal{DP}$, and the planner proceeds to the next part. %starting from the state where the last part $p^*$ has been disassembled. 
The sequence planner in PDP iteratively applies the path planner for each assembled part until either all parts have been disassembled or a timeout is reached. To limit the BFS search tree growth over obstructed parts that cannot be disassembled prior to other parts, the sequence planner restricts the maximum depth $d_{max}$ of a BFS tree.  The depth limit starts with $d_{max}=1$, and increments if no disassembled states have been found after searching for a disassembly path over all parts. We note that PDP reconstructs the entire search tree from scratch every time $d_{max}$ is incremented.

\textbf{Physics-Based Simulation}. In PDP, the motion trajectory is generated by the physics-based simulation that confines path planning to explore a physically valid subspace rather than the entire state space as in the sampling-based method \cite{lavalle1998rapidly}. Given a starting state $s$, an action $a$, and a candidate movement part $p$, the simulator iteratively applies $a$ over the part $p$ at state $s$ with a specified kinematic time step, producing a series of valid motion locations. The simulation ultimately returns either a reached disassembled state for part $p$ or a collision state when the distance between two generated locations is within the collision threshold.% \cite{tian2022assemble}. 

\textbf{Collision Detection}.
To simulate contact-rich disassembly tasks, the physics-based engine uses a Signed Distance Field (SDF) representation for each part. SDF enables accurate collision distance computation during trajectory generation. SDF $g_p(pt): \mathbb{R}^3 \rightarrow \mathbb{R}$ associates a point $pt \in \mathbb{R}^{3} $ with its closest distance to a part $p$,  where a negative value indicates that $pt$ is inside the geometry of $p$. SDF improves the collision distance computation for intricate assembly geometries, such as screws with fine threads \cite{tian2022assemble}.

\subsection{Directional Blocking Graph}

A \textit{Directional Blocking Graph} (DBG), introduced by \citet{wilson1992geometric}, defines a vertex as an individual part and edges as blocking relations given a specific movement direction.
%(Figure \ref{fig:DBG}).  
In a DBG$(d)$, an edge  $p_i\rightarrow p_j$ indicates that the part $p_j$ obstructs the translation of part $p_i$ along the direction $d$. Consequently, $p_i$ with an outdegree of zero, i.e. a sink vertex, indicates that a collision-free disassembly path exists for $p_i$ in direction $d$. DBGs can reduce the complexity of the disassembly planning problem as all precedence constraints between components are identified.  Different approaches have been introduced to build DBGs based on geometric and spatial analysis such as Minkowski differences 
\cite{lozano1993assembly,wilson1995two}.

\begin{figure}
    \centering
    % \captionsetup{font={small}}
    \includegraphics[width=0.8\textwidth]{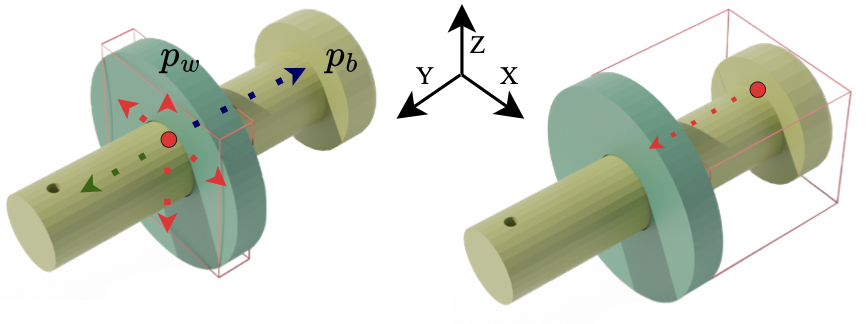}
            % \vspace{-0.5cm}
    \caption{An illustration of DBGs static analysis (left) with respect to a bolt $p_b$ and a washer $p_w$ along six directions;  a  demonstration of potential blocking assessment (right).}
    \label{fig:DBG}
\end{figure}

\begin{algorithm}[!ht]
\small
\SetAlgoLined
\KwIn{A disassembly problem $\mathcal{P}$ consisting of $m$ parts $P$ with the initial assembled state $s_0$, timeout $t_{\max}$, translational actions $A_t$, rotational actions $A_r$. }
\KwOut{An ordered sequence of disassembly paths $\mathcal{DP} := \{\mathcal{T}_{p_1},\ldots, \mathcal{T}_{p_m}\}$.}
$\mathcal{DP}:=\mathcal{T}:= \Pi_t:= \Theta_t:= \Pi_r:= \Theta_r:=\emptyset$; $P_r:=P$\;
$\Pi_t:=\Pi_t \cup (s_0, P_r, \mathcal{T})$; $\Pi_r:=\Pi_r \cup (s_0, P_r, \mathcal{T})$\;
\If{$s_o$ $\mathrm{is}$ $\mathrm{the}$ $\mathrm{goal}$ $\mathrm{state}$}{\Return{$\mathcal{DP}$}}

\SetKwFunction{Disassembly}{$disassembly$}
\SetKwProg{Fn}{Function}{:}{}
\SetKw{KwTo}{to}
\SetKwFor{For}{for}{do}{end for}
    \Fn{\Disassembly{$\Pi, \Theta, A$}}{
    \While{$\Pi \neq \emptyset$}{
    $s,P^*, \mathcal{T}:=extractNode(\Pi)$\;
    $\Theta:=\Theta \cup (s,P^*, \mathcal{T})$\;
    \If{$\mathrm{time}$ $t > t_{\max}$}{\Return{$\Pi,\Theta,\mathrm{true},\emptyset$}}
    \For{$\mathrm{part}$ $p$ $\mathrm{in}$ $P^*$}  {
    \For{$\mathrm{action}$ $a$ $\mathrm{in}$ $A$}
    { $s^\prime, t(s, s^\prime, a, p)$:=$f(a, s, p)${\scriptsize\tcp*[r]{Simulator}}
        $\mathcal{T}_p:=\mathcal{T} \cup t(s, s^\prime, a, p)$\;
                
        \If{isDisassembledState($s^\prime,p,P_r$)}{ 
        $\mathcal{DP}:=\mathcal{DP} \cup \mathcal{T}_p$;
        $P_r:=P_r\setminus p$\;
        
        {\scriptsize{\tcp{Delete nodes where $P^* = \{p\}$}}}
         $Delete(p)$\;   
        \If{$isGoalState(P_r)$}{\Return{$\Pi,\Theta, \mathrm{false}, \mathcal{DP} $}}
        $\Pi:=\Pi \cup \Theta$; $rmd(\Pi)$; $\Theta:=\emptyset$\;
        {\scriptsize \tcp{Terminate for Rotation}}
        \If{isRotational}{ \Return{$\Pi, \Theta, \mathrm{false},\emptyset$}}
        }
        \Else {{\scriptsize{\tcp{The collision state}}}
        $\Theta:=\Theta \cup (s^\prime, \{p\}, \mathcal{T}_p)$\  
        }
        
    }
    }
    }
     $\Pi:=\Pi \cup \Theta$;$rmd(\Pi)$;$\Theta:=\emptyset$\;
     \Return{$\Pi, \Theta, \mathrm{false}, \emptyset$}
    }
    
    {\While{$\mathrm{true}$}{
    {\scriptsize \tcp{Translational disassembly planning}}
    $\Pi_t, \Theta_t, \mathrm{timeout}, \mathrm{goal}$$:=$\Disassembly{$\Pi_t, \Theta_t, A_t$}\;
    \lIf{$\mathrm{timeout}$}{\Return{$\mathrm{failed}$}}
    \lIf{$\mathrm{goal} \neq \emptyset$}{\Return{$\mathcal{DP}$}}
    \tcp{\scriptsize Rotational disassembly planning}
     $\Pi_r, \Theta_r,\mathrm{timeout}, \mathrm{goal}$$:=$\Disassembly{$\Pi_r, \Theta_r, A_r$}\;
    \lIf{$\mathrm{timeout}$}{\Return{$\mathrm{failed}$}}
    \lIf{$\mathrm{goal} \neq \emptyset $}{\Return{$\mathcal{DP}$}}

}}

\caption{SBDP}
\label{alg:loops}
\end{algorithm}

{

\begin{figure*}[ht]
    \centering
        \includegraphics[width=1\textwidth]{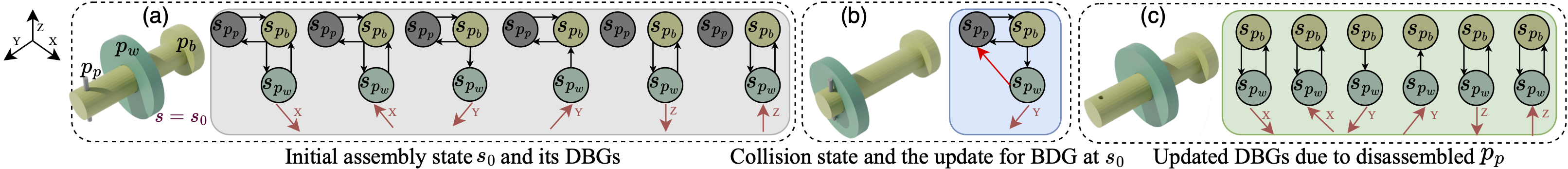}
        % \vspace{-0.7cm}
    \caption{DBGs examples with respect to construction through static analysis (grey) and updates due to the collision state (blue) and the disassembled state (green) in a problem $\mathcal{P}$ that consists of a bolt $p_b$, a washer $p_w$, and a pin $p_p$.}\label{fig:DBG_in_SBDP}

\end{figure*}
}
% \vspace{-8mm}
    
\section{State-Based Disassembly Planning}

State-Based Disassembly Planning (SBDP) approach stores the unexplored states in an \textit{open list} $\Pi$ and the resulting states, including visited states and simulated collision states, in a \textit{collision list} $\Theta$. 
%motion states that result in collisions to avoid the need to recompute computationally expensive simulations. 
SBDP explores disassembly paths of varying lengths by expanding stored states rather than a FIFO policy constrained with $d_{max}$ that needs to reconstruct the entire search tree every time $d_{max}$ is updated. SBDP expands one node at a time. When the initial state is expanded, it tries to disassemble all parts.  Otherwise, only part $p$ is enabled for disassembly when the expanded state  results from moving $p$. SBDP integrates the efficiency of translational motion and the coverage of rotational motion, by prioritizing disassembly planning with translational motion, denoted as translational disassembly planning, over disassembly planning with rotational motion, named as rotational disassembly planning. SBDP leverages translational disassembly planning to repeatedly disassemble each part until no further disassembly is feasible, and then applies rotational disassembly planning to disassemble remaining assemblies. To keep efficiency in SBDP,  rotational disassembly planning terminates upon the removal of one part, followed by translational disassembly planning to disassemble remaining assemblies. In SBDP, translational planning and rotational planning maintain the same disassembly plan $\mathcal{DP}$ and remaining assemblies, ensuring consistency in the parts requiring disassembly in either planning, and allowing $\mathcal{DP}$ to contain a mix of translational and rotational motions, through the concatenation of individual plans.  

Algorithm \ref{alg:loops} shows the pseudo-code of SBDP. The main loop, Lines 27-33, repeatedly executes translational disassembly planning (Line 28) and rotational disassembly planning (Line 31) with the rule that rotational planning is considered only after completing translational planning. SBDP implements  both translational and rotational planning within the disassembly function (Lines 5-26). Line 1 initializes the open list $\Pi$, collision list $\Theta$, disassembly plan $\mathcal{DP}$ and partial disassembly path $\mathcal{T}$ with the empty set. The open list $\Pi$ is updated with an unexplored node tuple  $\Pi=\{(s_0, P_r, \mathcal{T})\}$ (Line 2), where $P_r$ denotes remaining assemblies, initialized with $P$. We use subscripts $t$ and $r$ to denote $\Pi$, $\Theta$ and actions $A$ used in translational and rotational planning. SBDP returns an empty $\mathcal{DP}$ if the initial state is a goal state (Lines 3-4), or a disassembly plan $\mathcal{DP}$ on Line 30 or 33 when translational or rotational planning achieves the goal state  in the disassembly function (Lines 18-19); otherwise, a $failed$ is returned on Line 29 or 32 after reaching timeout in the disassembly function (Lines 9-10) for either planning approach.

The disassembly function receives the  open list $\Pi$, collision list $\Theta$, and disassembly actions $A$ from either planning approach as input and returns updated $\Pi$ and $\Theta$ on Line 26 after moving newly generated nodes from $\Theta$ to $\Pi$, removing duplicate nodes in $\Pi$ ($rmd(\Pi)$) and clearing $\Theta$ on Line 25. Lines 5-26 detail the implementation of the disassembly function. An unexplored node, containing a state $s$, candidate disassembly parts $P^*$, and a partial disassembly path $\mathcal{T}$, is selected and removed from the open list $\Pi$ on Line 7, where $P^*=P_r=P$ when $s=s_0$. The expanded node is inserted into the collision list $\Theta$ on Line 8 to record the visited state.  SBDP tries to disassemble a part $p \in P^*$ with an action $a \in A$, through the transition function $f(a, s, p)$ on Line 13. 
The transition function updates state $s$ to $s^\prime$ implemented by the physics-based simulation, and the simulated motion path $t$ from $s$ to $s^\prime$ is added to the partial disassembly path $\mathcal{T}$ (Line 14). If $s^\prime$ is a disassembled state for $p$, SBDP appends the disassembly path $\mathcal{T}_p$ to $\mathcal{DP}$, updates remaining assemblies $P_r$ with $P_r\setminus p$ (Line 16), and deletes nodes where $P^*=\{p\}$ from the open and collision lists of both planning approaches ($\Pi_t, \Pi_r, \Theta_t, \Theta_r$) (Line 17), as there is no need to find other disassembly paths for part $p$. For the same reason, SBDP updates $P^*$ with $P^*\setminus p$ for the node where $s=s_0$ in the \textit{Delete} function (Line 17). If $s^\prime$ is not the goal state, Line 20 transfers nodes from $\Theta$ to $\Pi$, eliminates duplicates in $\Pi$, and resets $\Theta$ to an empty set. This enables the continued search to reassess disassembly across all states in \textbf{translational planning}, since a disassembled part may unblock disassembly paths for other parts. We note that for \textbf{rotational planning}, SBDP terminates the disassembly (Lines 21-22) upon successfully disassembling one component to restart translational planning and reevaluate potential released blockages, as rotational planning is computationally demanding.  If $s^\prime$ is a collision state, SBDP inserts the node tuple ($s^\prime, \{p\}, \mathcal{T}_p$) into $\Theta$ (Line 24), resulting in only part $p$ being considered for disassembly when $s^\prime$ is expanded again. We note that for all other nodes where $s\neq s_0$, $P^*$ is set to $\{p\}$, with $p\in P$, as once a failed disassembly of part $p$ leads to a new node,  that node and all its child nodes commit to finding a disassembly path for $p$.
Disassembly function terminates, returns updated $\Pi$ and $\Theta$, and triggers the switch between translational and rotational planning when no part can be disassembled after searching for a disassembly path over all states in $\Pi$, leading to $\Pi=\emptyset$.

Compared with PDP, SBDP stores collision states to facilitate the disassembly path generation from the latest reached states rather than reconstructing the entire search tree from the initial state, which scales up the search significantly. To obtain a high-quality disassembly path, i.e. a short path,  while minimizing simulation computational cost, SBDP records visited states  but removes duplicate states. This allows SBDP to disassemble each part, starting from its initial state and all intermediate motion states.
SBDP achieves the goal state when remaining assemblies $P_r$ contain only two parts, and one of them is disassembled through either translational or rotational planning. We note that SBDP restricts the physics-based simulation with a time limit to prevent infinite simulation in both planning methods.

\section{DBGs in State-Based Disassembly Planning}

Different from previous work where DBGs are pre-computed prior to the search using geometric and spatial relations analysis, we introduce a method to build a DBG for each action (moving direction) based on the SDF representation, and a function to update DBGs using feedback from each simulation. In addition, we enrich DBGs in SBDP with state information, where SBDP constructs and updates DBGs for each state to capture the blockage relationships among different parts. For example, in a DBG$(a)$ associated with state $s$, its vertex is represented by  $s_p$, the location of a part $p$ at $s$, associated with a potential  blocking relationship via moving $p$ at $s$ along the direction defined by the planning action $a$. We introduce two novel evaluation functions derived from DBGs to guide SBDP. New evaluation functions enable SBDP to extract best-evaluated nodes from the open list in both translational and rotational planning on Line 7 in Algorithm 1, and avoid unnecessary simulations when collisions are detected and recorded in DBGs.

\begin{figure}[t]
    \centering
    \includegraphics[width=1\textwidth]{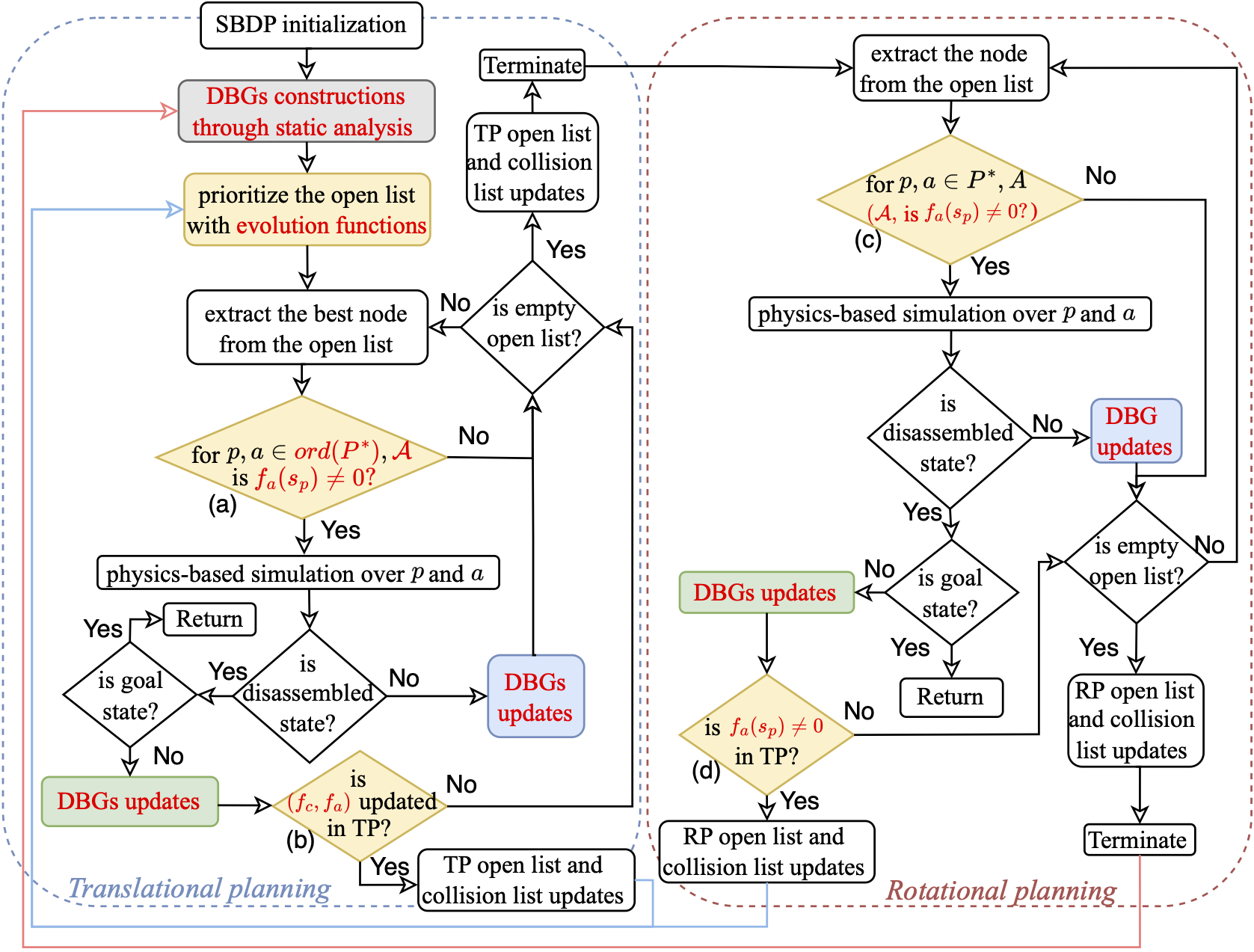}
    \caption{A flow chart of DBG-guided SBDP. DBGs constructions and updates are highlighted with the same colors used in Figure \ref{fig:DBG_in_SBDP} and usages are highlighted with yellow in both translational planning (left) and rotational planning (right). TP and RP are acronyms of translational planning and rotational planning, respectively. }
    \label{fig:flow_chart}
\end{figure}

\begin{figure*}[ht]
    \centering
     \includegraphics[width=1\textwidth]{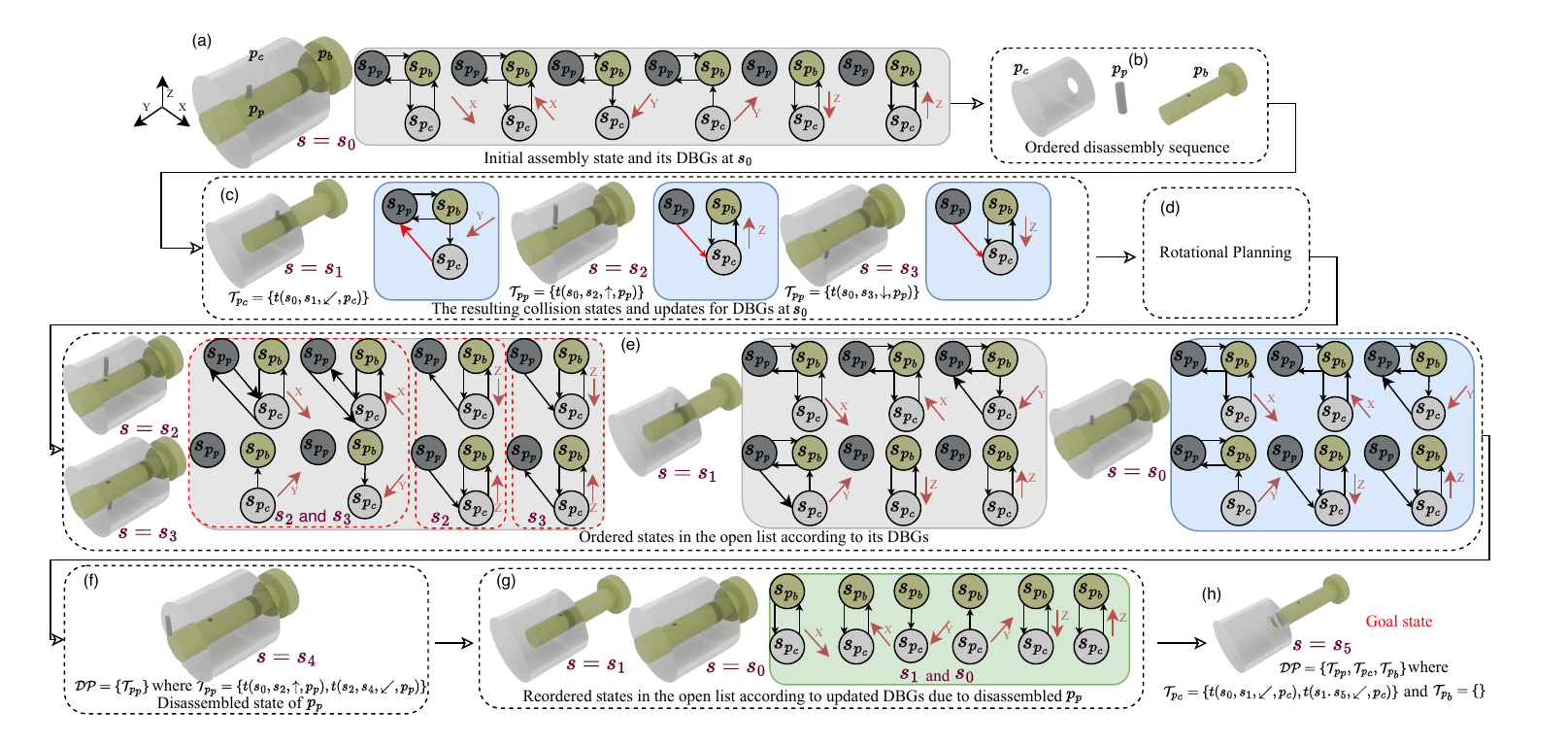}
    \caption{An example disassembly planning process with DBGs in SBDP for a problem $\mathcal{P}$ that consists of a bolt $p_b$, a cover $p_c$, and a pin $p_p$. The colored DBGs represent the same constructions and updates approaches as depicted in Figure \ref{fig:DBG_in_SBDP}.}
    \label{fig:DPexample}
\end{figure*}

\subsection{DBGs Static Analysis}
SBDP performs static analysis to build DBGs for each state. In static analysis, we consider all pairs of \textit{contacting} parts $(p_o, p_s)$  as potential blocking parts when an overlap exists between their collision shapes defined by axis-aligned minimum bounding boxes given  their locations $(s_{p_o}, s_{p_s})$ at state $s$. To establish the blocking relations between $p_o$ and $p_s$, one part, $p_o$, is designated as mobile, while the other, $p_s$, is considered stationary. We denote the mesh vertices of part $p_o$ within the overlapping area as contacting points $C_{p_o}$ and extend each contacting point $c_{p_o} \in C_{p_o}$ in disassembly directions over a specified number of steps resulting in a series of extension points $E_{p_o}$. If there is an extension point $e_{p_o} \in E_{p_o}$ whose SDF distance $g_{p_s}(e_{p_o}) < 0$, then a blocking relationship is built with an edge $s_{p_o}$ $\rightarrow$ $s_{p_s}$ added in the DBG of the extended direction and an edge $s_{p_s}$ $\rightarrow$ $s_{p_o}$ added in the DBG of the opposite direction. For example in the left part of Figure \ref{fig:DBG}, the red lines delineate the overlapping collision shape between the mobile bolt $p_b$ and the stationary washer $p_w$. A contacting point in $p_b$ and its six extension directions are shown, where the red arrow indicates a detected collision with $p_w$, and the blue arrow denotes the correct disassembly direction for $p_b$.

In certain situations, detecting blocking relations based solely on contacting points may be insufficient. For example, the bolt in Figure \ref{fig:DBG} is blocked along the positive Y direction, indicated with green, due to the bolt head. To address such scenarios, a further blocking assessment is considered, if the collision shape of  mobile part $p_o$ is larger than the collision shape of stationary part $p_s$ along the moving axis.  We first identify the potential collision area where the collision shape of $p_o$ lies behind the collision shape of $p_s$ in the movement direction. We then denote the mesh vertices of $p_o$ within the potential collision area as potential contacting points and repeat the same extension process as mentioned above to detect collisions. For example, in the right part of Figure \ref{fig:DBG}, we illustrate a potential contacting point in the red-outlined potential collision area, along its extension direction resulting in a blocking relationship.

In SBDP, static analysis is computed for the initial state and again for each generated collision state. To keep static analysis scalable, we leverage an adaptive extension distance in each extension step, which is a function of the length of the mobile part along the moving axis. Figure \ref{fig:DBG_in_SBDP} (a) displays a problem $\mathcal{P}$ that consists of a bolt $p_b$, a washer $p_w$, and a pin $p_p$, and its six DBGs generated through static analysis at $s_0$ with respect to six translational actions.  Static analysis allows SBDP to construct new DBGs for each state and further avoid the simulations when collisions have already been detected and recorded in the constructed DBGs. However, static analysis may overlook certain collisions due to the constraints on contacting pairs and the number of extension steps for the (potential) contacting point. This will be addressed and complemented in the DBGs updates function.

\subsection{DBGs Updates} 
DBGs undergo updates during motion planning according to the result of the transition function computed by physical simulation. If the transition function $f(s,a,p)$ leads to a collision state by moving part $p$ at state $s$ with action $a$, SBDP updates DBG$(a)$ at $s$ by adding edges from $s_p$ to the colliding parts.  Figure \ref{fig:DBG_in_SBDP} (b) illustrates a simulation outcome where the movement of $p_w$ at $s_0$ along the positive-Y direction ($\swarrow$) leads to a collision between $p_p$ and $p_w$, and the according DBG update, adding an arc $s_{p_w} \rightarrow s_{p_p}$ in the DBG($\swarrow$) at $s_0$. In contrast, if $f(s,a,p)$ results in a disassembled state, all DBGs in SBDP are updated, where all vertices and edges related to  $p$ are removed, indicating the release of all precedence constraints associated with $p$. For example, Figure \ref{fig:DBG_in_SBDP} (c) depicts DBGs, resulting from deleting all vertices and edges related to $p_p$, when $p_p$ is disassembled. We note that DBGs blockage updates are asymmetric since trajectories generated by physical simulation differ for colliding components along the opposite movement direction. For instance in Figure \ref{fig:DBG_in_SBDP} (a), the symmetric simulation over $p_p$ along the negative-Y direction would not lead to the collision with $p_w$ due to the initial obstruction with $p_b$. 

\subsection{DBGs Heuristics}

We introduce two evaluation functions in SBDP given a state $s$ and its DBGs: $f_a(s_{p})$ is the number of DBGs where $s_{p}$ is a sink vertex, i.e. the number of available collision-free actions for part $p$ at the location in state $s$; $f_c(s_{p})$ is the number of parts $p'\neq p$ pointed by the outgoing arcs of $s_{p}$ over all DBGs associated with state $s$, i.e. the number of colliding parts with part $p$ at the location in state $s$. The combination of $(f_c$, $f_a)$ is used to prioritize the open list,  using $f_c$ first, breaking ties with $f_a$. Specifically, the $(f_c$, $f_a)$ values for the node $(s, \{p\}, \mathcal{T}_p)$ are computed only for the part $p$ at state $s$, and the $(f_c$, $f_a)$ values for the node $(s_0, P^*, \mathcal{T})$ are set to infinite. Besides, $(f_c$, $f_a)$ is used to order the disassembly sequence on Line 11 in Algorithm \ref{alg:loops}, resulting in an ordered $P^*$, when $s_0$ is the expanded state. $f_c$ and $f_a$ are treated as heuristics, so smaller values are preferred. This encourages SBDP to explore a part with fewer obstructing components and fewer disassembly available actions, as having more disassembly actions would lead to more simulation computation via longer feasible trajectories.

\subsection{DBGs Constraints}

DBGs generated by static analysis and blockage updates enable SBDP to avoid unnecessary physics-based simulations over collision-detected parts with inapplicable actions. For a part $p$ at the expanded state $s$, SBDP allows the simulation to be computed with an action $a$ if $a$ is a collision-free action for part $p$, represented by the sink vertex $s_p$ in the DBG$(a)$ associated with state $s$. This constraint reduces the number of applicable actions $A$ on Line 12 in Algorithm \ref{alg:loops}, where we use $\mathcal{A}$  to denote satisfied actions. Furthermore, SBDP skips all simulations over part $p$ at the expanded state $s$ if no collision-free actions are available for part $p$, indicated by $f_a(s_p)=0$. This constraint prunes the duplicate state produced from the visited state in both translational and rotational planning. For instance, blockage updates for DBGs at an expanded state $s$ record all collisions as a result of disassembly attempts of part $p$, resulting in $f_a(s_p)=0$. Consequently, all simulations are skipped due to $f_a(s_p)=0$ when $s$ is expanded again. We note that the value of $f_a(s_p)$ will be updated through DBGs updates with respect to a disassembled component.

\section{DBG-Guided Disassembly Planning}

In SBDP, translational and rotational search states maintain their separate DBGs, but successful disassembly of a part updates all DBGs. Static analysis is only computed for translational planning, while it is omitted in rotational planning due to the computationally intense nature of rotation analysis. DBGs are set to empty whenever static analysis is required in rotational planning. However, rotational blockages are computed lazily while searching through DBGs updates.

The flowchart in Figure \ref{fig:flow_chart} describes DBGs constructions, updates and usages in SBDP,  where some SBDP specifics are eliminated to save space. SBDP initiates translational planning with static analysis for the initial state to construct its DBGs. 
Subsequently, the open list is prioritized based on $(f_c, f_a)$.  For each node extracted from the prioritized open list, translational planning performs the simulation when $f_a(s_p)\neq 0$  over the ordered $P^*$ and constrained translational actions $\mathcal{A}$ (decision (a)). For each collision state, updated DBGs record the simulated blockages to avoid redundant simulations. If a disassembled state leads to $(f_c, f_a)$ value updates, i.e. a new collision-free action, for states in the open or collision list (decision (b)), translational planning continues the search with an updated open list (Line 20 in Algorithm \ref{alg:loops}) that is prioritized again according to each state's new DBGs (blue flowline). For rotational planning, the absence of static analysis results in an exhaustive search over candidate disassembly parts $P^*$ and rotational actions $A$ (decision (c)). However, the updated blockages in DBGs allow rotational planning to focus on constrained actions $\mathcal{A}$ and simulations where $f_a(s_p)\neq0$ when exploring visited states. Rotational planning reactivates translational planning to reorder its open list while skipping static analysis (blue flowline), when a disassembled state results in any collision-free actions, indicated by $f_a(s_p) \neq 0$, for at least one state in the translational planning open list (decision (d)). Otherwise, SBDP restarts translational planning to build DBGs for each generated collision state in its open list through static analysis after the completion of rotational planning (red flowline).

DBGs enable an informed SBDP with a reduced search space. The recorded blockage relationships in DBGs inform SBDP to reassess disassembly in translational planning only for states whose DBGs contain  collision-free actions, whether existing or newly generated by the removal of a component in either translational or rotational planning.

Figure \ref{fig:DPexample} illustrates a SBDP disassembly planning process guided by DBGs for a problem $\mathcal{P}$ with three components: a pin $p_p$,  a bolt $p_b$ and a cover $p_c$. SBDP starts translational planning from the initial state $s_0$ as shown in part (a) where DBGs associated with $s_0$ generated through static analysis overlook the blockages between $p_c$ and $p_p$ due to the limited extension steps. Translational planning dequeues $s_0$, candidate disassembly parts $P^*=\{p_p, p_b, p_c\}$, and partial disassembly path $\mathcal{T}=\emptyset$ from its open list $\Pi_t$. The explored state  $s_0$ is then added to the collision list, updating it to $\Theta_t = \{(s_0, P^*, \mathcal{T})\}$. Translational planning  disassembles over ordered $P^*$ with the simulation restriction, $f_a(s_p)\neq 0$, and constrained translational actions $\mathcal{A}$ where $p_c$ is prioritized first, followed by $p_p$ and $p_b$ (part (b)) based on the evaluation functions $(f_c(s_{p_c})=1, f_a(s_{p_c})=1)$ for $p_c$,  $(f_c(s_{p_p})=1, f_a(s_{p_p})=2)$ for $p_p$ and $(f_c(s_{p_b})=2, f_a(s_{p_b})=0)$ for $p_b$, where $s=s_0$.

Part (c) depicts the generated collision states, recorded in the collision list $\Theta_t$, and corresponding blockage updates with respect to DBGs at $s_0$, where each collision state results from its simulated motion path $t$. The state $s_1$ describes the collision between $p_c$ and $p_p$ via the simulation over $p_c$ along  the only applicable action, positive-Y direction, in which the simulation leads to the path $t(s_0,s_1,\swarrow,p_c)$; $s_2$ and $s_3$ represent the collisions between $p_p$ and $p_c$ along the positive-Z ($\uparrow$) and negative-Z ($\downarrow$) movements over part $p_p$, resulting in the paths $t(s_0,s_2,\uparrow,p_p)$ and $t(s_0,s_3,\downarrow,p_p)$, respectively. Added arcs for DBGs at $s_0$ are highlighted and the simulation over $p_b$ is skipped due to $f_a(s_{p_b})=0$. For each produced collision state, translational planning updates the partial disassembly path $\mathcal{T}$ with the motion path $t$. Translational planning terminates disassembly since no parts are disassembled, resulting in $\Pi_t=\emptyset$.  It then renews $\Pi_t$ with $\Theta_t$ (Line 25 in Algorithm \ref{alg:loops}), leading to  $\Pi_t=\{(s_1, \{p_c\}, \mathcal{T}_{p_c}),(s_2,\{p_p\}, \mathcal{T}_{p_p}),(s_3,\{p_p\},\mathcal{T}_{p_p}),(s_0, P^*, $ $ \mathcal{T})\}$, where $\mathcal{T}$ is empty and $\mathcal{T}_{p_p}$ records the different $t$ for states $s_2$ and $s_3$. Subsequently, rotational planning is activated, while it finally reaches the same collision states (part d) and reactivates translational planning again due to $\Pi_r=\emptyset$.

Part (e) demonstrates the prioritized open list $\Pi_t=\{s_2,s_3,s_1,s_0\}$ in translational planning based on DBGs for collision states $s_2$, $s_3$ and $s_1$ generated through static analysis and DBGs for $s_0$ generated through blockage updates detailed in part (c). We only show the state configuration for each node to save space and highlight similarities and differences between DBGs for $s_2$ and $s_3$.
From left to right, $s_2$ and $s_3$ share the same priority due to the same ($f_c(s_{p_p})=1, f_a(s_{p_p})=3$) values, followed by $s_1$ with ($f_c(s_{p_c})=2, f_a(s_{p_c})=0$) values, and then the initial state $s_0$ with ($f_c=\infty, f_a=\infty$) values.
Translational planning randomly selects either $s_2$ or $s_3$ as the expanded state. If $s_2$ is chosen, the simulation begins from this state, moving $p_p$ along the positive-Y, negative-Y, or negative-Z direction. The positive-Y movement ultimately reaches the disassembled state $s_4$ with the motion path $t(s_2,s_4,\swarrow,p_p)$, resulting in a disassembly path $\mathcal{T}_{p_p}=\{t(s_0,s_2,\uparrow,p_p), t(s_2,s_4,\swarrow,p_p)\}$ for $p_p$ that is added to the disassembly plan $\mathcal{DP}$ as depicted in part (f).

The disassembled $p_p$ raises DBGs updates, deleting all vertices and edges related to $p_p$ in DBGs at $s_1$ and $s_0$, and continues translational planning with an updated $\Pi_t$ (Line 20 in Algorithm \ref{alg:loops}) as shown in part (g) where $s_2$ and $s_3$ are excluded, and $p_p$ is deleted from $s_0$ (Line 17 in Algorithm \ref{alg:loops}). We note that deleting updates result in the same DBGs for $s_1$ and $s_0$, while $s_1$ is prioritized over $s_0$ in $\Pi_t$ since the $(f_c, f_a)$ values for $s_0$ consistently remain infinite during prioritization. Part (h) shows the disassembled state for part $p_c$ accomplished through the simulation starting from the expanded state $s_1$ over part $p_c$ along the positive-Y direction. Subsequently, SBDP terminates due to goal achievement and returns the disassembly plan $\mathcal{DP}$, including the correct disassembly sequence for each part and its corresponding disassembly path, as the solution.

\begin{table}[t]
\small
% \captionsetup{font={small}}
\setlength{\tabcolsep}{5.2pt}
\renewcommand{\arraystretch}{1.2}
\begin{tabular}{c|c|ccc||c}
& \begin{tabular}[c]{@{}c@{}}All \\ (4196)\end{tabular} & \begin{tabular}[c]{@{}c@{}}Small \\ (2620)\end{tabular} & \begin{tabular}[c]{@{}c@{}}Medium \\ (1469)\end{tabular} & \begin{tabular}[c]{@{}c@{}}Large \\ (107)\end{tabular} & \begin{tabular}[c]{@{}c@{}} Difference \\ with SBDP$^*$
\end{tabular} \\ \hline
SBDP$^*$ & \textbf{4135}                                                  & \textbf{2603}                                                    & \textbf{1444}                                                     & \textbf{88}                                                     & -0/+0   \\
SBDP     & 4106                                                  & 2599                                                    & 1425                                                     & 82                                                     & -36/+7  \\
PDP$^*$  & 3976                                                  & 2551                                                    & 1348                                                     & 77                                                     & -160/+1 \\
PDP$_r$  & 3961                                                  & 2551                                                    & 1344                                                     & 66                                                     & -175/+1 \\
PDP$_t$  & 3772                                                  & 2460                                                    & 1247                                                     & 65                                                     & -363/+0
\end{tabular}
% \vspace{-0.3cm}
\caption{The number of disassembly problems solved by SBDP$^*$, SBDP, PDP$^*$, PDP$_r$, and PDP$_t$ in the small, medium, and large categories; The last column describes the number of problems solved by SBDP$^*$ while remaining unsolved by the planner in each row vs. problems solved by  the planner in each row  while remaining unsolved by SBDP$^*$.}
\label{table_solved}

\end{table}
\begin{table*}[t]
\small
\centering
% \captionsetup{font={small}}
\setlength{\tabcolsep}{2.1pt}
\renewcommand{\arraystretch}{1.2}
\begin{tabular}{c|cccccc|cccccc|cccccc|cccccc}
         & \multicolumn{6}{c|}{Small (2550)}                                                            & \multicolumn{6}{c|}{Medium (1339)}                                                            & \multicolumn{6}{c|}{Large (65)}                                                                & \multicolumn{6}{c}{All (3954)}                                                               \\ \hline
         & \multicolumn{2}{c|}{Sim}    & \multicolumn{2}{c|}{PT(s)}     & \multicolumn{2}{c|}{DT(s)} & \multicolumn{2}{c|}{Sim}     & \multicolumn{2}{c|}{PT(s)}     & \multicolumn{2}{c|}{DT(s)} & \multicolumn{2}{c|}{Sim}     & \multicolumn{2}{c|}{PT(s)}       & \multicolumn{2}{c|}{DT(s)} & \multicolumn{2}{c|}{Sim}     & \multicolumn{2}{c|}{PT(s)}     & \multicolumn{2}{c}{DT(s)} \\
         & Avg & \multicolumn{1}{c|}{Std} & Avg & \multicolumn{1}{c|}{Std} & Avg        & Std           & Avg & \multicolumn{1}{c|}{Std}  & Avg & \multicolumn{1}{c|}{Std} & Avg          & Std         & Avg & \multicolumn{1}{c|}{Std}  & Avg  & \multicolumn{1}{c|}{Std}  & Avg          & Std         & Avg & \multicolumn{1}{c|}{Std}  & Avg & \multicolumn{1}{c|}{Std} & Avg         & Std         \\ \hline
SBDP$^*$ & \textbf{12}  & \multicolumn{1}{c|}{\textbf{72}}  & \textbf{17}  & \multicolumn{1}{c|}{171} & \textbf{68}         & 187           & \textbf{55}  & \multicolumn{1}{c|}{\textbf{212}}  & \textbf{83}  & \multicolumn{1}{c|}{\textbf{220}} & \textbf{163}          & \textbf{283}         & \textbf{177} & \multicolumn{1}{c|}{\textbf{228}}  & \textbf{493}  & \multicolumn{1}{c|}{\textbf{539}}  & \textbf{703}          & \textbf{639}         & \textbf{29}  & \multicolumn{1}{c|}{\textbf{142}}  & \textbf{47}  & \multicolumn{1}{c|}{\textbf{210}} & \textbf{111}         & \textbf{254}         \\
SBDP     & 26  & \multicolumn{1}{c|}{171} & 24  & \multicolumn{1}{c|}{204} & 71         & 218           & 125 & \multicolumn{1}{c|}{366}  & 149 & \multicolumn{1}{c|}{443} & 219          & 542         & 373 & \multicolumn{1}{c|}{526}  & 729  & \multicolumn{1}{c|}{771}  & 883          & 823         & 65  & \multicolumn{1}{c|}{269}  & 78  & \multicolumn{1}{c|}{337} & 134         & 394         \\
PDP$^*$  & 71  & \multicolumn{1}{c|}{1K}  & 27  & \multicolumn{1}{c|}{212} & 87         & 248           & 312 & \multicolumn{1}{c|}{2.8K} & 185 & \multicolumn{1}{c|}{476} & 266          & 519         & 472 & \multicolumn{1}{c|}{875}  & 1K   & \multicolumn{1}{c|}{1.3K} & 1.2K         & 1.4K        & 159 & \multicolumn{1}{c|}{1.8K} & 97  & \multicolumn{1}{c|}{394} & 166         & 435         \\
PDP$_r$      & 25  & \multicolumn{1}{c|}{151} & 27  & \multicolumn{1}{c|}{\textbf{101}} & 88         & \textbf{158}       & 249 & \multicolumn{1}{c|}{789}  & 335 & \multicolumn{1}{c|}{695} & 417          & 740         & 825 & \multicolumn{1}{c|}{2.5K} & 1.6K & \multicolumn{1}{c|}{1.8K} & 1.7K         & 1.8K        & 114 & \multicolumn{1}{c|}{588}  & 157 & \multicolumn{1}{c|}{524} & 226         & 564        
\end{tabular}
% \vspace{-0.3cm}
\caption{ Comparisons of SBDP$^*$, SBDP, PDP$^*$, and PDP$_r$ in terms of avg. and std-dev. number of simulation executions (Sim), path planning time (PT), and dissasembly planning time (DT) across \textbf{commonly} solved problems. K is $10^3$. Best results are in bold. See Table \ref{table:comparsion} in the Appendix for additional comparisons.}
\label{table:detail_information}
\end{table*}

\begin{figure}[t]
    \centering
    \includegraphics[width=1\textwidth]{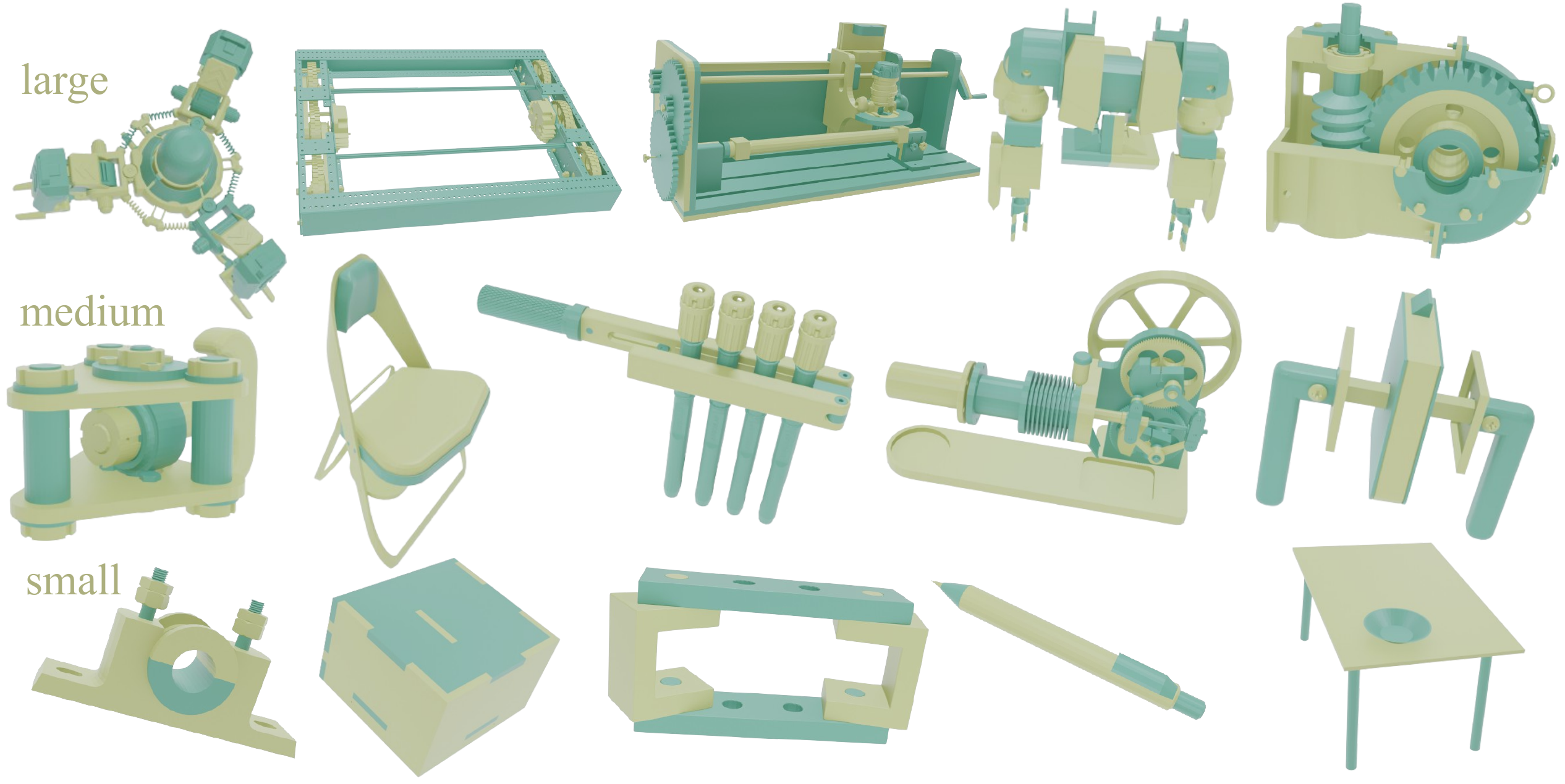}
    % \vspace{-0.7cm}
    \caption{Disassembly benchmark examples from the small, medium, and large categories.}
    \label{fig:examples}
\end{figure}

\section{Evaluation}

\citet{tian2022assemble} introduced a large-scale dataset comprising thousands of physically valid industrial assemblies, with comprehensive geometric pre-processed data for collision detection. We catergorize assemblies consisting of 3-9 (small), 10-49 (medium), and 50$+$ (large) components as disassembly benchmarks, with 4196 problems in total.  Figure \ref{fig:examples} illustrates disassembly examples from each category. We denote PDP$_t$ and PDP$_r$ as PDP with translational or rotational motion, respectively. PDP$_t$, PDP$_r$ and the variant PDP$^*$ serve as baselines, where PDP$^*$ leverages PDP$_t$ to disassemble all parts, and if unsuccessful, it uses PDP$_r$ to disassemble again. We denote SBDP$^*$ as DBG-guided SBDP with $(f_c, f_a)$ and DBGs constraints. All experiments were conducted on a cloud computer with clock speeds of 2.00 GHz Xeon processors and processes time out after 2 hours. We note that in PDP$^*$, both translational and rotational search procedures are allotted 2 hours each. The hyper-parameter settings for all methods are available in Table \ref{tab:hp_new} in the Appendix.

Table \ref{table_solved} compares the number of problems solved by different disassembly planners. SBDP$^*$ surpasses all other planners in each category and solves most of the problems, 4135 out of 4196 problems, accounting for 98.55\% of the total.  SBDP$^*$ and SBDP consistently outperform PDP$^*$, PDP$_r$  and PDP$_t$ showcasing the effectiveness of state-based search in disassembly planning. Utilizing DBGs information, SBDP$^*$ successfully addresses 36 problems unsolved by SBDP, emphasizing the significance of DBGs. Compared with SBDP$^*$, SBDP solves extra 7 problems, and PDP$^*$ and PDP$_r$ each solves a unique problem unsolved by SBDP$^*$. In these problems, we empirically observed occasional incorrect collision detection, resulting in penetration disassembly trajectories for all approaches.

Table \ref{table:detail_information} compares SBDP$^*$, SBDP, PDP$^*$, and PDP$_r$ in terms of search efficiency, including path planning time (PT) and disassembly planning time (DT), as well as search space, indicated by the number of executed simulations (Sim) across commonly solved problems. Due to the limited problem coverage from Table \ref{table_solved}, PDP$_t$ is excluded. We note that path planning time only counts the simulation cost while disassembly planning time considers the overall cost, i.e. static analysis, DBGs updates, and path planning in SBDP$^*$. In PDP$^*$, only successful planning time and executed simulations are recorded, excluding failures in translational or rotational search. SBDP$^*$ consistently outperforms all other planners, with the lowest average and standard deviation search space, and the lowest average planning time, achieving reductions of 55\%/80\%/75\% in average search space, 40\%/50\%/70\% in average path planning time, and 20\%/35\%/50\% in average disassembly planning time compared to SBDP, PDP$^*$, and PDP$_r$, across all problems. In the medium and large groups, SBDP$^*$ exhibits notable efficiency by reducing the average search space, path planning time, and disassembly planning time by 55\%, 45\%, and 25\% compared to SBDP, and by 70\%, 55\%, and 40\% compared to the best performance of PDP$^*$ and PDP$_r$. SBDP maintains the same advantages as SBDP$^*$ in terms of search space and planning time compared to the best performance of PDP$^*$ and PDP$_r$, reducing the average search space by 45\%, path planning time by 20\%, and disassembly planning time by 20\% in total, and demonstrates the same outstanding performance in the medium and large groups. 

\begin{figure}[t]
    \centering
        % \captionsetup{font={small}}
    \includegraphics[width=1\textwidth]{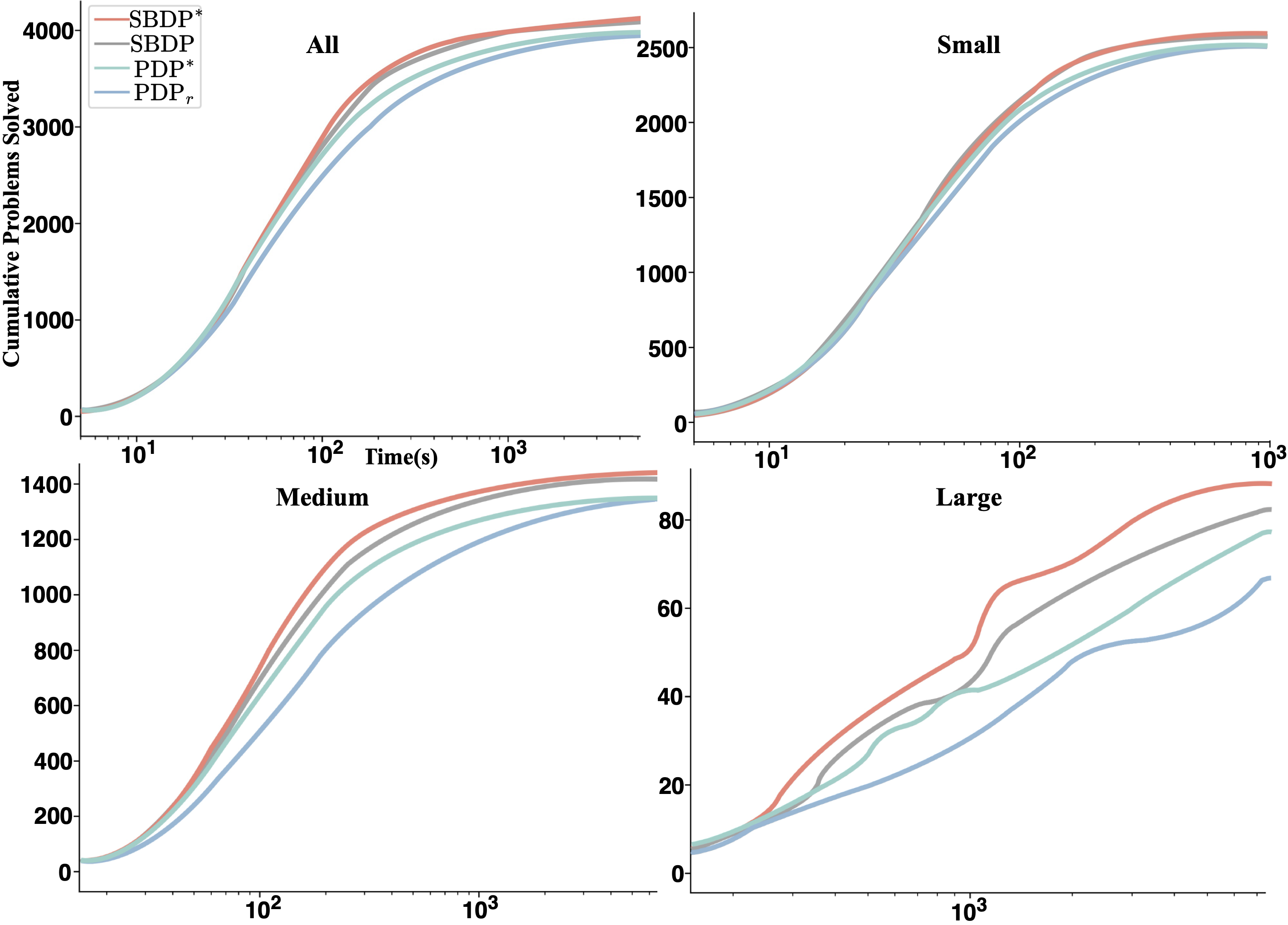}
    % \vspace{-0.7cm}
    \caption{Coverage of solved problems as a function of time for planners SBDP$^*$, SBDP, PDP$^*$ and PDP$_r$ across total problems and each assembly size category. See Figures \ref{fig:line_all}--\ref{fig:line_large} in the Appendix for coverage graphs of commonly solved problems}
    \label{fig:covered_problem}
\end{figure}

The curves in Figure \ref{fig:covered_problem} display the number of problems solved as a function of time for different planners. SBDP$^*$ and SBDP solve more problems than PDP$^*$ and PDP$_r$ even for short time windows. This is especially evident in the medium and large assembly categories. This trend aligns with the results from Table \ref{table_solved} and Table \ref{table:detail_information}, highlighting the benefits of state-based search and DBGs. SBDP$^*$ demonstrates state-of-the-art performance in terms of success rate and computational efficiency.

\section{Discussion}

By using static analysis and updates on DBGs with state information, DBG-guided state-based disassembly planning consistently outperforms the current state-of-the-art disassembly planner. SBDP benefits from state-based search and DBG-derived evaluation functions.  However, applying mature planning techniques, such as classical planners with search heuristics \cite{Hoffmann_2001,helmert2006fast,lei2021width}, in SBDP remains challenging. In SBDP, effects become available only after physical simulation, with no declarative representation available. Recent studies on leveraging \textit{novelty}-based algorithms over Functional STRIPS representation for solving planning problems with simulators and numerical features, such as classical control problems \cite{lipovetzky2021planning}, provide potential improvement approaches. 
DBGs information is non-trivial for complex disassembly problems with numerous components, as the intricate blockages among different parts constrain the disassembly sequence and path. As a result, a comprehensive analysis of precedence constraints via DBGs is necessary to determine the available disassembly strategy, which is missed in PDP. We note that our DBGs initialization concentrates solely on contacting parts and uses SDF for efficient collision detection, completing within seconds.

The state-based disassembly planning can be readily applied to assembly-by-disassembly tasks by reversing the disassembly sequence and paths and connecting them with the initial states in the assembly task to construct the entire assembly sequence and paths.  For each component, a collision-free path from its initial state in the assembly task to its disassembled state in the disassembly task can be easily generated through motion planning algorithms such as RRT-Connect~\cite{kuffner2000rrt} given the existence of fewer motion constraints between these two states \cite{tian2022assemble}. The assembly path can be obtained through connecting the generated path with the reversed disassembly path. %By connecting the generated path with the reversed disassembly path, we can obtain the assembly path for each component, 
Subsequently, by reverse looping over the disassembly sequence and repeating the same assembly path generation method, we can ultimately derive an ordered assembly sequence containing all assembly paths.

\section{Conclusion}

We introduce the state-based search paradigm for disassembly planning (SBDP) which avoids redundant simulations of explored states. SBDP integrates both translational and rotational motions without requiring user input to choose between them. We show that new evaluation functions derived from DBGs improve the success rate and efficiency of SBDP across assemblies with varying number of components. DBG-guided SBDP demonstrates state-of-the-art performance in disassembly planning.  In the future, other learning-based methods \cite{zakka2020form2fit, tian2023asap} can be incorporated to guide SBDP. Additionally, disassembly plans generated by SBDP can be employed in assembly training systems \cite{zhu2018robot} to facilitate robot learning through demonstrations \cite{thomas2018learning, fan2019learning, de2021autonomous}. 

\section*{Acknowledgements}

Chao Lei is supported by Melbourne Research Scholarship established by The University of Melbourne.

This research was supported by use of The University of Melbourne Research Cloud, a collaborative Australian research platform supported by the National Collaborative Research Infrastructure Strategy (NCRIS).

\bibliography{aaai25}
\clearpage
\appendix

\section{Appendix}

\begin{figure}[hp]
    \centering
    \includegraphics[width=1\textwidth]{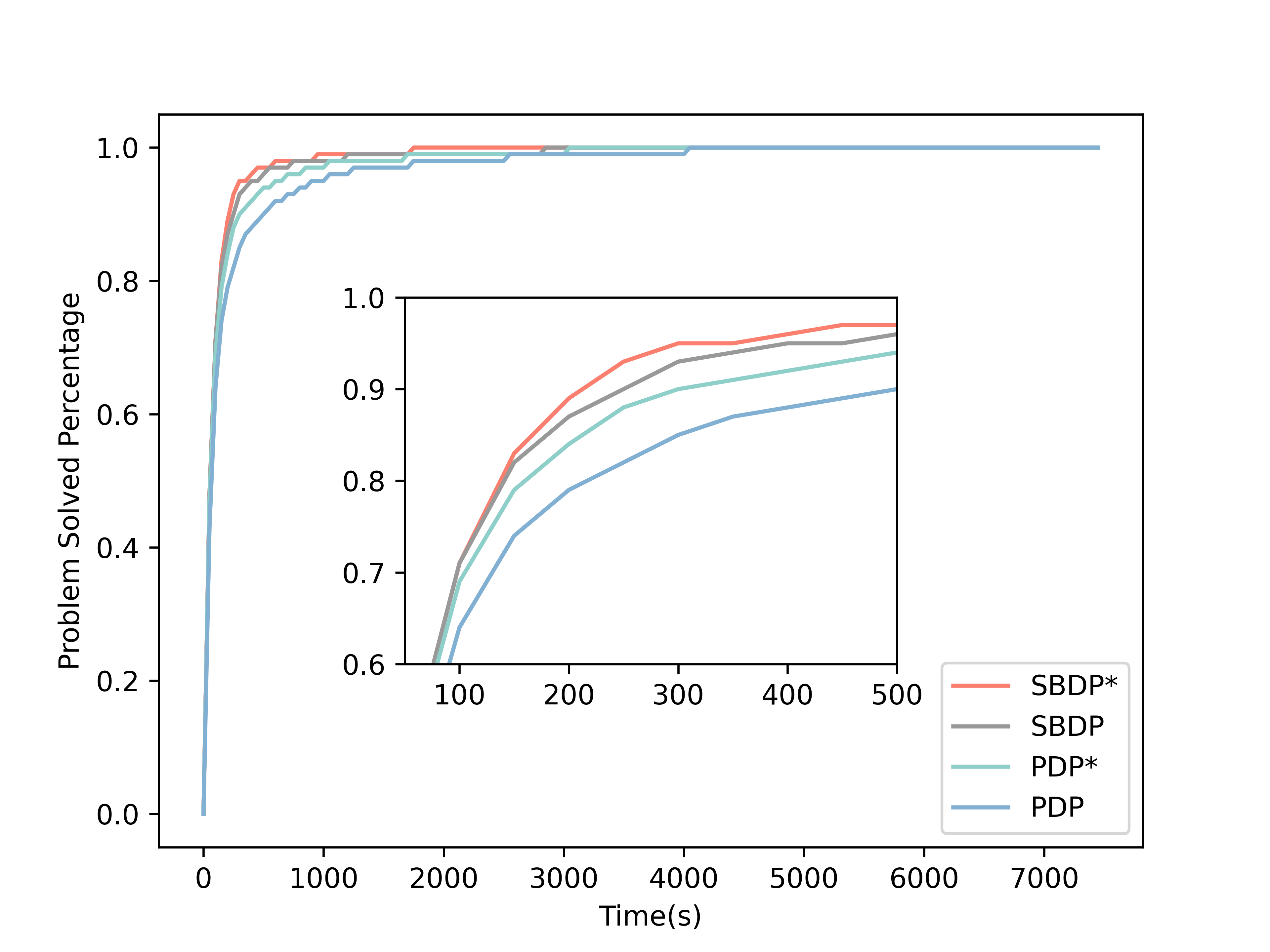}
    \caption{Coverage percentage of commonly solved problems \textbf{in total} as a function of time for different disassembly planners.}
    \vspace{-2mm}
    \label{fig:line_all}
\end{figure}

\begin{figure}[hp]
    \centering
    \includegraphics[width=1\textwidth]{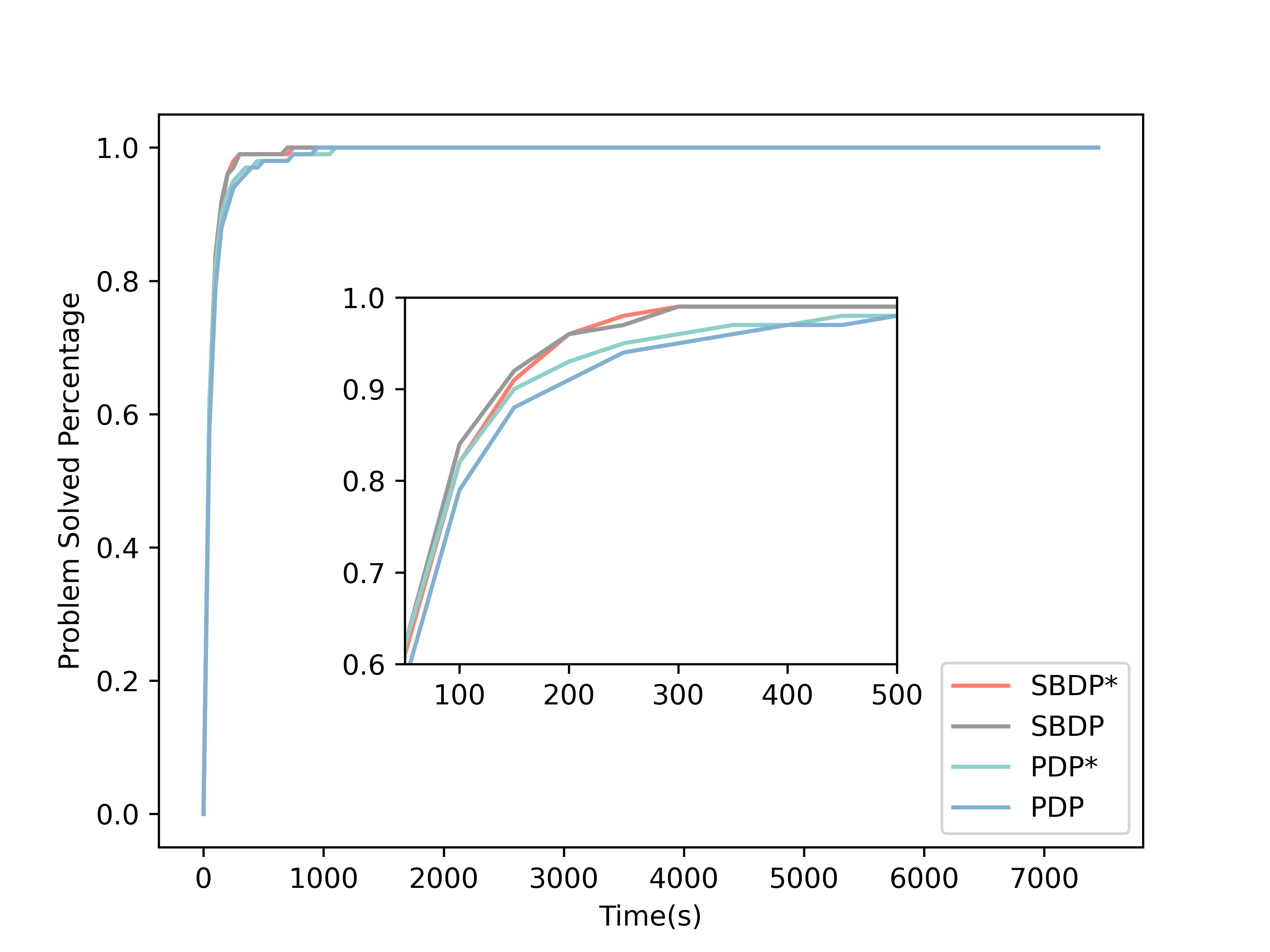}
    \caption{Coverage percentage of commonly solved problems in the \textbf{small} group as a function of time for different disassembly planners.}
    \vspace{-2mm}
    \label{fig:line_small}
\end{figure}

\begin{figure}[bp]
    \centering
    \includegraphics[width=1\textwidth]{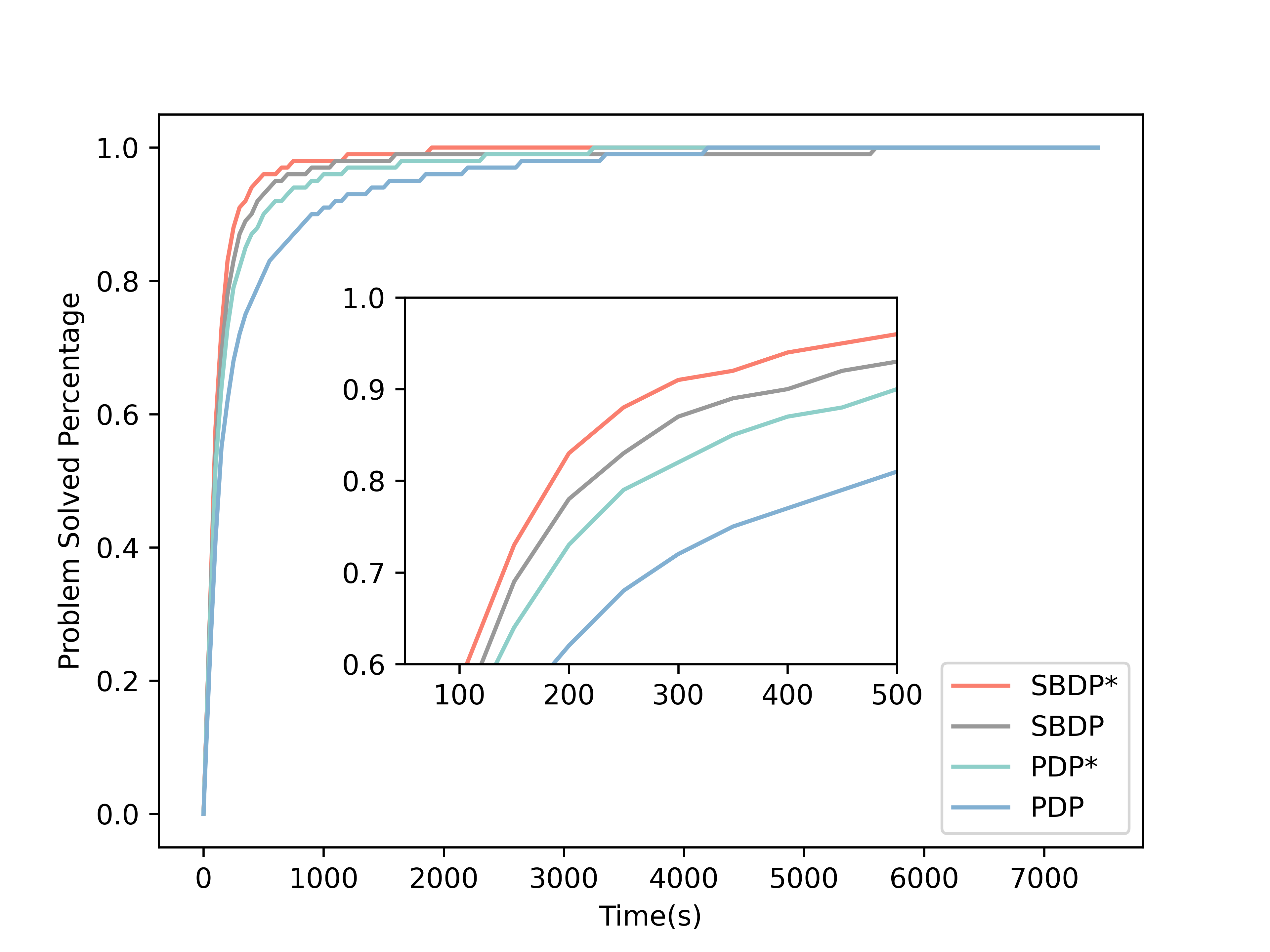}
    \caption{Coverage percentage of commonly solved problems in the \textbf{medium} group as a function of time for different disassembly planners.}
    \vspace{-2mm}
    \label{fig:line_medium}
\end{figure}

\begin{figure}[bp]
    \centering
    \includegraphics[width=1\textwidth]{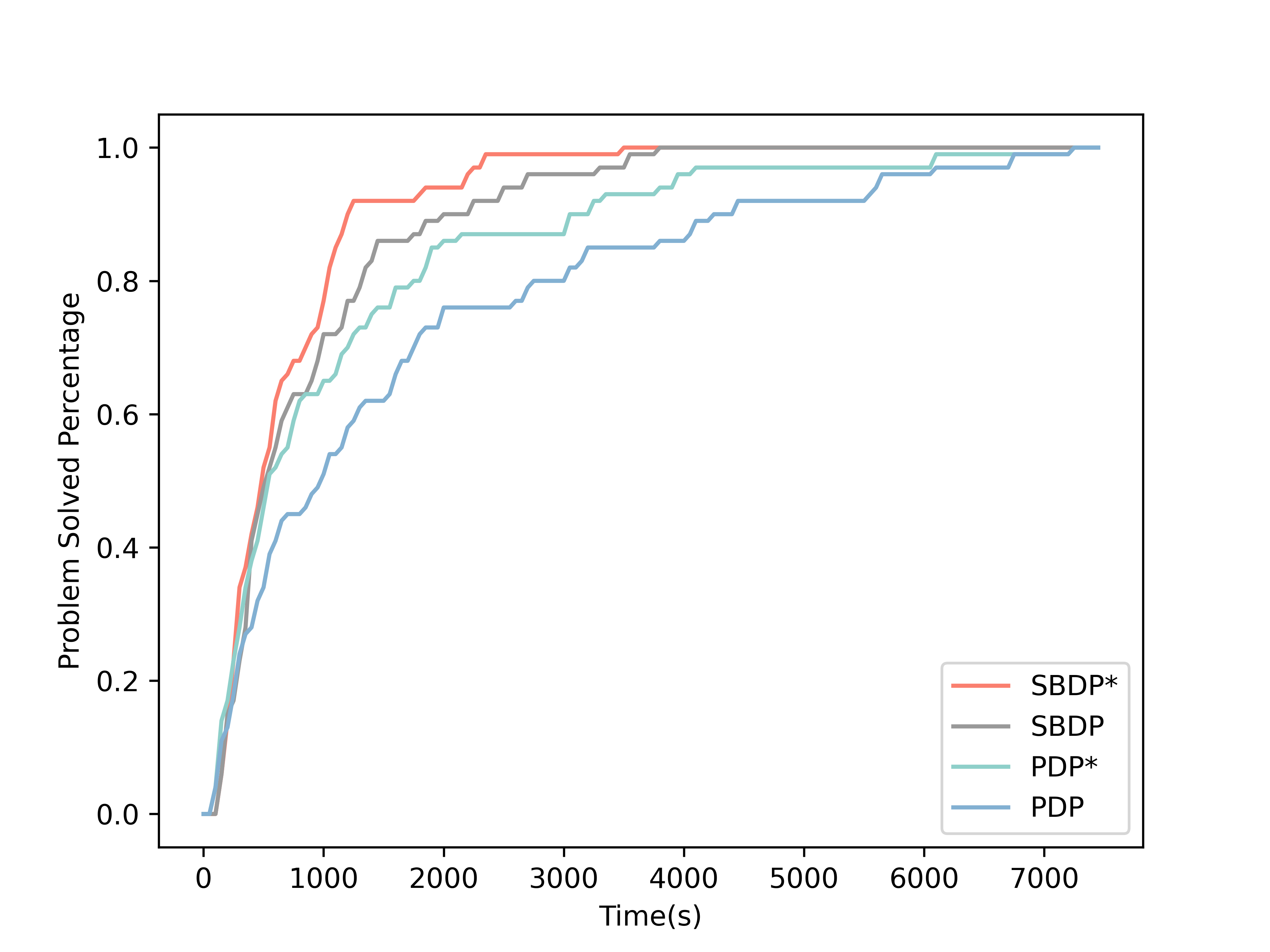}
    \caption{Coverage percentage of commonly solved problems in the \textbf{large} group as a function of time for different disassembly planners.}
    \vspace{-2mm}
    \label{fig:line_large}
\end{figure}

\begin{table*}[!ht]
\begin{center}
\begin{tabular}{l|lc}
                                      & Name                                                & Value                \\ \hline
\multirow{8}{*}{Disassembly Planning} & \textbf{Number of extension steps}                           & 5                   \\
                                      & \textbf{Adaptive extension distance}                         & $\min(0.05, L_i/20)$ \\
                                      & Path planning timeout                               & 360s                 \\
                                      & Path planning time step                  & 1e-1                 \\
                                      & Penetration threshold for collision detection       & 0.01                 \\
                                      & Force/torque magnitude of each action               & 100                  \\
                                      & State similarity threshold (translation) $\delta_t$ & 0.05                 \\
                                      & State similarity threshold (rotation) $\delta_r$    & 0.5                  \\ \hline
\multirow{3}{*}{Simulation}           & Contact stiffness $k_n$                             & 1e6                  \\
                                      & Contact damping coefficient $k_d$                   & 0                    \\
                                      & Simulation time step                                & 1e-3                
\end{tabular}
\caption{Hyper-parameters for SBDP, SBDP$^*$, PDP$_t$, PDP$_r$, and PDP$^*$. 
The highlighted parameters are exclusive to SBDP$^*$. $L_i$ represents the length of the component along the $i$-th dimension, assuming each assembly is scaled to fit within a 10x10x10 unit cube.}
\label{tab:hp_new}
\end{center}
\end{table*}

\begin{table*}[h]
\centering
\footnotesize
\setlength{\tabcolsep}{1.3pt}
\renewcommand{\arraystretch}{1.2}
\begin{tabular}{c|cccc|cccc|cccc|cccc}
            & \multicolumn{4}{c|}{\begin{tabular}[c]{@{}c@{}}Small\\ (2550)\end{tabular}} & \multicolumn{4}{c|}{\begin{tabular}[c]{@{}c@{}}Medium\\ (1339)\end{tabular}} & \multicolumn{4}{c|}{\begin{tabular}[c]{@{}c@{}}Large\\ (65)\end{tabular}} & \multicolumn{4}{c}{\begin{tabular}[c]{@{}c@{}}All\\ (3954)\end{tabular}} \\ \hline
            & \multicolumn{2}{c|}{PP}                         & \multicolumn{2}{c|}{DP}   & \multicolumn{2}{c|}{PP}                         & \multicolumn{2}{c|}{DP}    & \multicolumn{2}{c|}{PP}                        & \multicolumn{2}{c|}{DP}  & \multicolumn{2}{c|}{PP}                       & \multicolumn{2}{c}{DP}   \\
            & LR        & \multicolumn{1}{c|}{ TR/TI}      & LR        &  TR/TI     & LR        & \multicolumn{1}{c|}{ TR/TI}      & LR        &  TR/TI      & LR       & \multicolumn{1}{c|}{ TR/TI}      & LR       &  TR/TI     & LR       & \multicolumn{1}{c|}{ TR/TI}     & LR       &  TR/TI     \\ \hline
SBDP$^*$/PDP  & 68\%      & \multicolumn{1}{c|}{-21/+18}        & 49\%      & -48/+17        & 83\%      & \multicolumn{1}{c|}{-280/+38}         & 70\%      & -338/+35       & 77\%     & \multicolumn{1}{c|}{-1320/+126}     & 71\%     & -1405/+151    & 73\%     & \multicolumn{1}{c|}{-156/+25}       & 56\%     & -239/+22      \\
SBDP$^*$/PDP$^*$ & 67\%      & \multicolumn{1}{c|}{-25/+22}        & 45\%      & -53/+17       & 79\%      & \multicolumn{1}{c|}{-123/+36}       & 60\%      & -166/+32       & 74\%     & \multicolumn{1}{c|}{-716/+189}      & 62\%     & -841/+168     & 71\%     & \multicolumn{1}{c|}{-78/+29}       & 50\%     & -147/+23     
\end{tabular}
\caption{ Comparisons of SBDP$^*$ with PDP and PDP$^*$ with respect to  the percentage of problems where SBDP$^*$ with less time (LR) in path planning (PP) and disassembly planning (DP); average path planning time and disassembly planning time reduction/increase (TR/TI) in seconds for problems where SBDP$^*$ spends less/more time in path planning and disassembly planning.}
\label{table:comparsion}
\end{table*}

\end{document}